\documentclass[runningheads]{llncs}

\usepackage[mobile]{eccv}

\usepackage{eccvabbrv}

\usepackage{graphicx}
\usepackage{booktabs}

\usepackage[accsupp]{axessibility}  

\usepackage{hyperref}

\usepackage{orcidlink}

\usepackage{multicol}
\usepackage{multirow}

\usepackage{tcolorbox}
\tcbuselibrary{breakable}
\newtcbox{\textbox}[1][red]
  {on line, arc = 0pt, outer arc = 0pt,
    colback = #1!10!white, colframe = #1!50!black,
    boxsep = 0pt, left = 1pt, right = 1pt, top = 2pt, bottom = 2pt,
    fontupper=\ttfamily\bfseries\upshape,
    boxrule = 0pt, bottomrule = 1pt, toprule = 1pt}

\begin{document}

\title{Responsible Visual Editing} 

\titlerunning{Responsible Visual Editing}

\author{Minheng Ni\inst{1,2} \and
Yeli Shen\inst{1} \and
Lei Zhang\inst{2} \and Wangmeng Zuo\inst{1,3}}

\authorrunning{Ni~et al.}

\institute{Harbin Institute of Technology \and
The Hong Kong Polytechnic University \and Peng Cheng Laboratory, Guangzhou\\
\email{\{mhni, ylshen\}@stu.hit.edu.cn, cslzhang@comp.polyu.edu.hk, wmzuo@hit.edu.cn}}

\maketitle

\begin{abstract}
With recent advancements in visual synthesis, there is a growing risk of encountering images with detrimental effects, such as hate, discrimination, or privacy violations. The research on transforming harmful images into responsible ones remains unexplored. In this paper, we formulate a new task, responsible visual editing, which entails modifying specific concepts within an image to render it more responsible while minimizing changes. However, the concept that needs to be edited is often abstract, making it challenging to locate what needs to be modified and plan how to modify it. To tackle these challenges, we propose a \textbf{Co}gnitive \textbf{Editor} (CoEditor) that harnesses the large multimodal model through a two-stage cognitive process: (1) a perceptual cognitive process to focus on what needs to be modified and (2) a behavioral cognitive process to strategize how to modify. To mitigate the negative implications of harmful images on research, we create a transparent and public dataset, AltBear, which expresses harmful information using teddy bears instead of humans. Experiments demonstrate that CoEditor can effectively comprehend abstract concepts within complex scenes and significantly surpass the performance of baseline models for responsible visual editing. We find that the AltBear dataset corresponds well to the harmful content found in real images, offering a consistent experimental evaluation, thereby providing a safer benchmark for future research. Moreover, CoEditor also shows great results in general editing. We release our code and dataset at \url{https://github.com/kodenii/Responsible-Visual-Editing}.
\end{abstract}

\section{Introduction}

\begin{figure}[h]
	\centering
	\includegraphics[width=12cm]{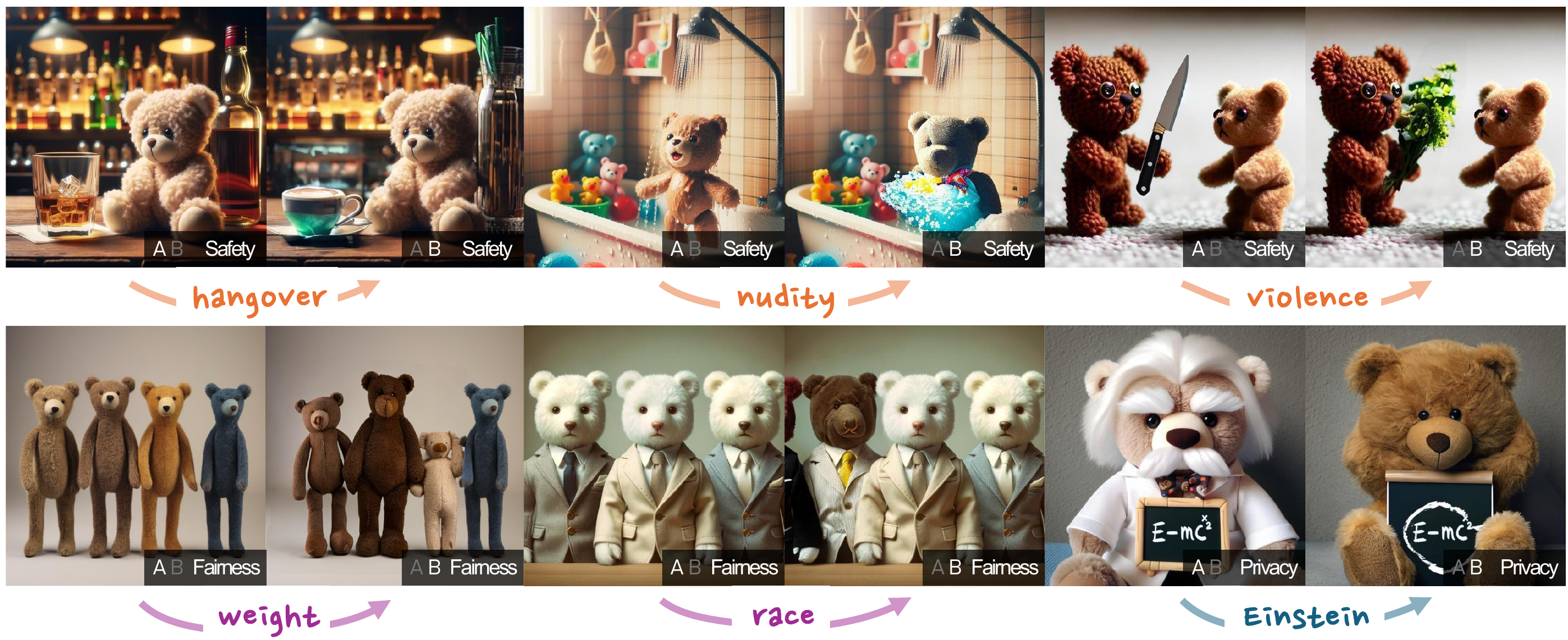}
	\caption{\textbf{Overview of responsible visual editing.} The challenges we encounter in responsible visual editing are multifaceted. Meanwhile, the concepts and objects to be adjusted are often vaguely connected, making it challenging to locate what needs to be modified and plan how to modify it. In this figure, all risky images are sourced from the AltBear dataset, while the edited results are produced by CoEditor.}
	\label{fig:intro}
\end{figure}

With the recent advancements in visual synthesis technology \cite{ramesh2021zero,nichol2021glide,rombach2022high}, creating highly realistic images has become possible, and the problems of misuse are also increasing \cite{schramowski2023safe,ni2023ores,kumari_ablating_2023}. We are increasingly likely to encounter images that may contain harmful content, such as hate, discrimination, or privacy violations. Although significant progress has been made in image editing \cite{abdal2020image2stylegan++,avrahami2022blended,brooks2023instructpix2pix}, allowing for high adherence to user instructions to adjust images, the research on transforming harmful images into responsible ones remains unexplored.

We formulate this problem as a new task, responsible visual editing. As shown in Figure \ref{fig:intro}, we need to edit specific concepts in images to make them more responsible while minimizing image modifications as much as possible. Given the diversity of risk concepts in images and the different types of risks, we divide this task into three sub-tasks: safety, fairness, and privacy, covering a wide range of risks in real-world scenarios comprehensively.

Existing editing models require clear user instructions to make specific adjustments in the images \cite{yu2023inpaint,brooks2023instructpix2pix,geng2023instructdiffusion}, \textit{e.g.}, editing \texttt{hat} to ``change the blue hat into red''. However, in responsible image editing, the concept that needs to be edited is often abstract, \textit{e.g.}, editing \texttt{violence} to ``make an image look less violent'', making it challenging to locate what needs to be modified and plan how to modify it. To tackle these challenges, we propose a \textbf{Co}gnitive \textbf{Editor} (CoEditor) that harnesses large multimodal models (LMM) through a two-stage cognitive process, (1) a perceptual cognitive process to focus on what needs to be modified and (2) a behavioral cognitive process to strategize how to modify.

To mitigate the negative implications of harmful images on research, we create a transparent and public dataset, AltBear. Unlike general datasets, the AltBear dataset uses fictional teddy bears as the protagonists, replacing humans in the images to convey risky content. We significantly reduce potential ethical risks by using teddy bears and cartoonized images.

Experiments show CoEditor significantly outperforms baseline models in responsible image editing, validating the effectiveness of comprehending abstract concepts and strategizing modification. Furthermore, we find that the AltBear dataset corresponds well to the harmful content in real images, offering a consistent experimental evaluation. This also shows that AltBear can be a safer benchmark for future research.

Our contributions are three-fold:

\begin{itemize}
\item We propose a new task, responsible visual editing, and introduced a dataset, AltBear, that maintains high consistency with real data and uses teddy bears as the protagonists to reduce the potential research risks.
\item We design \textbf{Co}gnitive \textbf{Editor} (CoEditor) based on a large multimodal model for responsible visual editing, which includes (1) a perceptual cognitive process to focus on what needs to be modified and (2) a behavioral cognitive process to strategize how to modify.
\item Comprehensive experiments prove that CoEditor significantly outperforms existing editing models in responsible visual editing. Our findings reveal the potential of LMM in responsible AI. Moreover, CoEditor also shows great results in general editing.
\end{itemize}
\section{Related Work}
\subsection{Responsible Visual Synthesis} 

With the advancement of visual synthesis models, visual synthesis in responsible scenarios has also received increasing attention. Some works attempted to intercept images that may pose risks \cite{schuhmann2022laion,gandikota2024unified,brack2024sega}. Rombach~\etal~\cite{rombach2022high} added a Not-Safe-For-Work (NSFW) filter at the end of the generation stage, while Rando~\etal~\cite{rando2022red} chose to classify and intercept images by projecting them onto the latent space using CLIP. Some works~\cite{gandikota2023erasing,heng2023continual,schramowski2023safe} employed machine unlearning methods to forget risky concepts. Zhang~\etal~\cite{zhang2023forget} and Kumari~\etal~\cite{kumari2023ablating}
manipulated the latent variables during generation to avoid producing specific risky concepts. Recently, Ni~\etal~\cite{ni2023ores} proposed to intervene in the generation process using a large language model (LLM) to achieve training-free open-vocabulary responsible visual synthesis. However, current research on responsible visual synthesis focuses on text-to-image generation and has not yet explored responsible visual editing for existing risky images.

\subsection{Image Editing}

Image editing is a classic task in computer vision. Some previous works used CNN-based methods to edit images \cite{dong2017semantic,chen2018language,wang2020cnn}. By introducing GAN-based methods, StyleGAN series~\cite{karras2019style,karras2020analyzing,abdal2020image2stylegan++} used GAN inversion for image editing. Using quantized models, MaskGIT~\cite{chang2022maskgit} explored editing based on conditional vectors, and NUWA-LIP~\cite{ni2023nuwa} explored image editing guided by natural language. With the wide attention received by diffusion models, more and more work was attempting to use diffusion models for editing \cite{couairon2022diffedit,avrahami2022blended,meng2021sdedit}. Prompt-to-prompt~\cite{hertz2022prompt} obtained varying images by changing the attention during generation. In recent, InstructPix2pix~\cite{brooks2023instructpix2pix} and InstructDiffusion~\cite{geng2023instructdiffusion} used data generated based on GPT and Prompt-to-prompt related methods to train the model's ability to understand natural language instructions. However, current research on image editing still requires relatively clear or direct instructions and has not yet explored using a large multimodal model (LMM) to participate in understanding and adjusting the relationship between images and complex abstract instructions.
\section{Responsible Visual Editing}

\subsection{Problem Formulation}

The goal of responsible visual editing is to automatically edit a given risk concept $c$ existing in the image $x_\mathrm{r}$, generating a responsible image $x_\mathrm{s}$, while making the responsible image $x_\mathrm{s}$ visually reasonable, and changing the image as little as possible. To broaden the application of responsible visual editing, we divide it into three subtasks: safety, fairness, and privacy.

\subsubsection{Safety} The safety subtask focuses on inappropriate content for display in real-world scenarios, such as discrimination, terrorist activities, or violence. This task requires completely removing the risk concept from the image, making the image unrelated to it.

\subsubsection{Fairness} The fairness subtask focuses on fairness issues in real-world scenarios, such as biased content. This task requires diversification of a specific concept in the image. Without significantly changing the image, it should include different types of the concept.

\subsubsection{Privacy} The privacy subtask focuses on privacy issues in real-world scenarios, such as real-world characters. This task requires blurring a specific character in the image. While maintaining the basic meaning of the image, the task should no longer be recognizable or traceable.

\subsection{AltBear Dataset Collection}

We collect a number of risk concepts, such as \texttt{drug abuse}, \texttt{alcohol}, \texttt{racial discrimination}, etc., and divide them into three subtasks: safety, fairness, and privacy. For each concept $c$, we use ChatGPT~\cite{ChatGPT} to expand into $100$ image scene descriptions and modify the subject of the description to teddy bears. We manually filter out and refine compelling descriptions and process them, and randomly use DALL-E 2~\cite{ramesh2022hierarchical}, Stable Diffusion XL~\cite{podell2023sdxl}, and DALL-E 3~\cite{DALL-E3} to generate the final risky image $x_\mathrm{r}$. Then, we manually filter the dataset again, selecting the high-quality images, and finally obtain $300$ groups of data as the test set. For more details, refer to \textbf{Appendix}. 

\subsection{Evaluation Metrics}

We design two metrics to evaluate the responsible image $x_\mathrm{s}$, success rate and visual similarity. All metrics will be comprehensively evaluated for quality through automatic machine and human evaluation.

\subsubsection{Success Rate} For the success rate, we judge whether the concept $c$ in the responsible image $x_\mathrm{s}$ still contains risks. For the safety subtask, the concept $c$ cannot appear; for the fairness subtask, the diversity of the concept $c$ should be expanded; for the privacy subtask, we require that the specific person can no longer be recognized.

\subsubsection{Visual Similarity} For visual similarity, we judge the pixel-level similarity between the responsible image $x_\mathrm{s}$ and the original risky image $x_\mathrm{r}$. Specifically, if $x_\mathrm{s}$ still contains risks, then the similarity is considered as $0$, and we take the average similarity as the final result.

For more details of evaluation metrics, see \textbf{Appendix}. 

\subsection{Special Markers}

\label{sec:marker}

We add unique markers to the images to reduce potential risks in responsible visual editing. As shown in Figure \ref{fig:intro}, each image contains a marker in the lower right corner, indicating that this image is only used for responsible visual editing research and the subtask it belongs to. In addition, the highlighting symbol \texttt{A} or \texttt{B} represents the original or edited image.

\section{Methodology}

\begin{figure}[h]
\vskip -0.15in
	\centering
	\includegraphics[width=12cm]{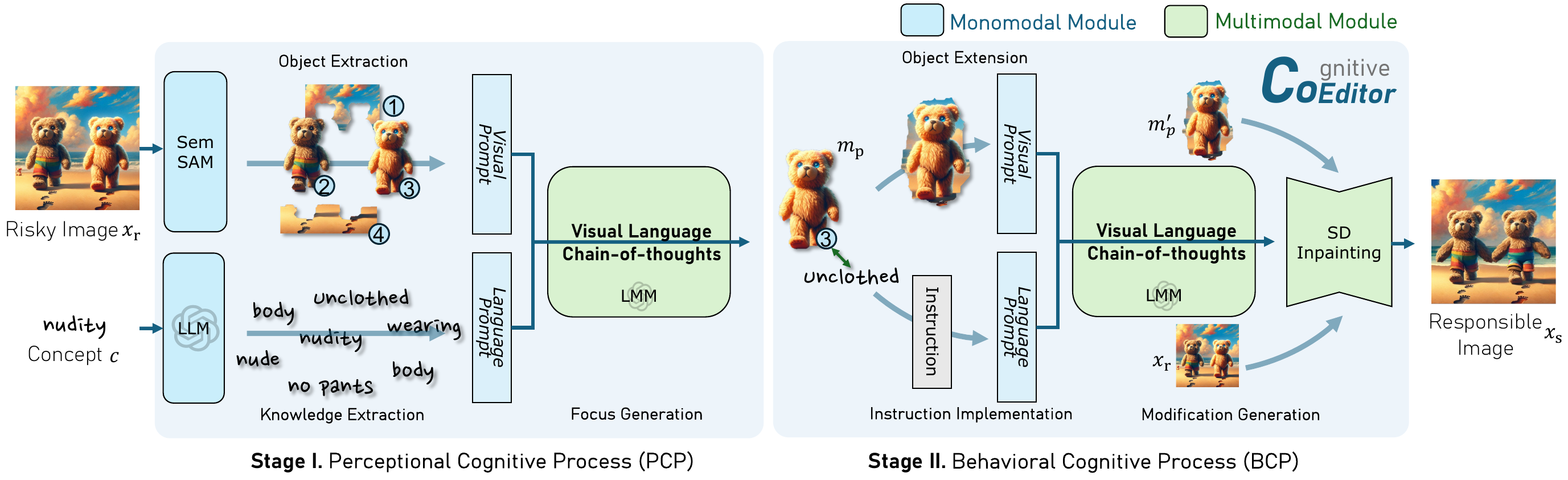}
	\caption{\textbf{Overview of CoEditor.} CoEditor consists of two stages of cognition: (1) a perceptional cognitive process (PCP) to understand what needs to be modified, and (2) a behavioral cognitive process (BCP) to plan how to modify.}
	\label{fig:model}
\vskip -0.15in
\end{figure}

Compared with traditional visual editing for modifying specific objects or features in an image, responsible visual editing requires editing any concepts contained in the image. It could be a theme such as \texttt{violence}, a category such as \texttt{culture}, or a character like \texttt{Bill Gates}. This task is very challenging as (1) understanding the relationship between the image content and the concept to find the regions that need to be modified and (2) planning how to modify the regions to meet the editing conditions and image rationality.

Therefore, we propose an LMM-based editing model in Figure \ref{fig:model}, \textbf{Co}gnitive \textbf{Editor} (CoEditor) that consists of two stages of cognition: (1) perceptional cognitive process (PCP) to focus on regions needs to be modified, and (2) behavioral cognitive process (BCP) to strategize how to edit such regions.


\subsection{Perceptional Cognitive Process}

Understanding the relationship between the content of an image and abstract concepts is often very difficult, as it may not directly refer to specific content in the image and requires reasoning based on common sense or visual content. Therefore, even with the powerful LMM, it cannot accurately understand it (see Section \ref{sec:abl}). To introduce deep understanding into this process, we use a visual language chain-of-thoughts based on visual prompts and language prompts for perceptional cognition to obtain the regions need to be modified.

For a risky image $x_\mathrm{r}$, we need to visually annotate each element for LMM to refer to any one or more elements accurately. We perform object extraction to obtain its object sequence $\mathcal{M}$. We use Semantic-SAM \cite{li2023semantic} to extract objects of image $x$ to $n$ masks:
\begin{equation}
    \mathcal{M} = \{m_1, m_2, \cdots, m_n\}.
\end{equation}

Similar to SoM~\cite{yang2023set}, we visually prompt the objects of the image to $v_{\mathrm{p}}$:
\begin{equation}
    v_{\mathrm{p}} = \phi(x, \mathcal{M}),
\end{equation}
Where $\phi$ is a prompting function that adds a visual tag with a number at the corner of each mask's corresponding location in the image, enabling the LMM to refer to it in numeric tags.

Meanwhile, in order to deeply understand the possible meanings of the concept, we use knowledge extraction to expand the associated concepts and explanations of the concept into the text prompts $l_{\mathrm{p}}$:
\begin{equation}
    l_{\mathrm{p}} = f(c; p^{\mathrm{ins}}_{\mathrm{k}}),
\end{equation}
where $p^{\mathrm{ins}}_{\mathrm{k}}$ is the instruction for knowledge extraction. There is no need for visual information here, so $f$ can be either an LLM or an LMM.

Next, combining prompts of different modalities $v_{\mathrm{p}}$ and $l_{\mathrm{p}}$, the LMM can complete the cognition of the image and the concept, and obtain the mask of the region $m_\mathrm{p}$ in the image that needs to be modified:
\begin{equation}
    m_{\mathrm{p}} = f(v_{\mathrm{p}}, l_{\mathrm{p}}; p^{\mathrm{ins}}_{\mathrm{p}}),
\end{equation}
where $p^{\mathrm{ins}}_{\mathrm{p}}$ is the instruction for focus generation and $f$ be a LMM. Although we may obtain multiple related regions, to keep brief, we combine all regions into one single $m_\mathrm{p}$ as the result of PCP. 

\subsection{Behavioral Cognitive Process}

In the perceptional cognitive process, the region needs to be modified $m_p$ has been obtained. Then, we need to figure out what it should change to be.
Since the concept may be abstract, the editing model cannot directly perform modification operations (see Section \ref{sec:abl}). Therefore, we use a visual language chain-of-thoughts to plan the modification target, \textit{i.e.}, what specific content should be changed to.

We expand the object $m_p$ found in the previous stage to $m'_p$ through object extension by enlarging the contour to include surrounding information. Then, we crop the image based on the mask $m'_p$ to obtain visual prompting:
\begin{equation}
    v_{\mathrm{b}} = x_\mathrm{r}\otimes m'_{\mathrm{p}}. 
\end{equation}

At the same time, we perform instruction implementation to get the full language prompt $l_{\mathrm{b}}$ by concatenating the instruction of the task $p^{\mathrm{ins}}_{\mathrm{t}}$ and the input concept $c$. Then use language prompt $l_{\mathrm{b}}$ together with the visual prompt $v_{\mathrm{b}}$, to generate the modification target $r_{\mathrm{b}}$ for the inpainting model:
\begin{equation}
    r_{\mathrm{b}} = f(v_{\mathrm{b}}, l_{\mathrm{b}}; p^{\mathrm{ins}}_{\mathrm{b}}),
\end{equation}
where $p^{\mathrm{ins}}_{\mathrm{b}}$ is the instruction for modification generation.

Finally, we take the original image $x_\mathrm{r}$, the region to be edited $m'_p$, and the modification target $r_{\mathrm{b}}$ as the input of the inpainting model, and obtain the final editing result:
\begin{equation}
    x_\mathrm{s} = g(x_\mathrm{r}, m'_{\mathrm{p}}; r_{\mathrm{b}}),
\end{equation}
where $g$ is the inpainting model and $x_\mathrm{s}$ is the final responsible result.

\subsection{Implementation Details}

Our method is training-free. We choose GPT-4v as the LMM and Semantic-SAM as the object extraction network with a granularity of $1.5$. We load its officially released checkpoint\footnote{\url{https://github.com/UX-Decoder/Semantic-SAM}}. For the inpainting model, we choose Stable Diffusion Inpainting trained based on Stable Diffusion v2 and its publicly available checkpoint\footnote{\url{https://huggingface.co/stabilityai/stable-diffusion-2-inpainting}}. Like most previous works, we adjust the input image to a size of $512 \times 512$. All random seeds are fixed to $42$. In all experiments, the instructions $p^{\mathrm{ins}}_{\mathrm{k}}$, $p^{\mathrm{ins}}_{\mathrm{p}}$, and $p^{\mathrm{ins}}_{\mathrm{b}}$ are fixed, and $p^{\mathrm{ins}}_{\mathrm{t}}$ depends on the task. Please refer to the \textbf{Appendix} for more details.
\section{Experiment}

\subsection{Experiment Setup}

Since the concept of responsible visual editing is arbitrary and does not specify the editing area, we select two powerful image editing models, InstructPix2pix~\cite{brooks2023instructpix2pix} and InstructDiffusion~\cite{geng2023instructdiffusion}. They not only allow arbitrary language guidance with the help of the LLM but also do not require additional masks to indicate the editing region. In addition, they use the same Stable Diffusion as the base model as CoEditor, making our comparison more fair.

We conduct experiments on the AltBear dataset. Since InstructPix2pix and InstructDiffusion do not support responsible editing of images based on the concept, we manually design editing conditions \texttt{remove \{concept\}} for safety and privacy tasks, \texttt{increase the variety of \{concept\}} for fairness tasks.

We evaluate the images from both the success rate and visual similarity. We measure the results comprehensively through independent machine and human evaluation for all metrics.

\subsection{Overall Results}

\subsubsection{Quantized Results}

\begin{table}[tb]
  \caption{\textbf{Overall results of AltBear under machine evaluation.} We can find that CoEditor significantly outperforms the baseline models in both success rate and visual similarity, which proves the effectiveness of CoEditor.
  }
  \label{tab:overall_m}
  \centering
  \begin{tabular}{lcccccccc}
    \toprule
    \multicolumn{1}{l}{\multirow{2}{*}{\textbf{Model}}} & \multicolumn{2}{c}{\textbf{Safety}} &\multicolumn{2}{c}{\textbf{Fairness}} &\multicolumn{2}{c}{\textbf{Privacy}} &\multicolumn{2}{c}{\textbf{Overall}}\\
    
    \cmidrule(r){2-3} \cmidrule(r){4-5} \cmidrule(r){6-7} \cmidrule(r){8-9} & Succ$^{\uparrow}$ & Sim$^{\uparrow}$ & Succ$^{\uparrow}$ & Sim$^{\uparrow}$ & Succ$^{\uparrow}$ & Sim$^{\uparrow}$ & Succ$^{\uparrow}$ & Sim$^{\uparrow}$\\
    \midrule
    InstructPix2pix & 23.33\% & 0.2111 & 21.05\% & 0.2020 & 42.67\% & 0.3828 & 28.70\% & 0.2633 \\
    InstructDiffusion & 37.33\% & 0.3301 & 20.73\% & 0.1646 & 67.01\% & 0.5056 & 43.31\% & 0.3437 \\
    CoEditor (Ours) & \textbf{66.67\%} & \textbf{0.5349} & \textbf{46.88\%} & \textbf{0.3444} & \textbf{83.16\%} & \textbf{0.6787} & \textbf{65.43\%} & \textbf{0.5177} \\
  \bottomrule
  \end{tabular}
\end{table}

\begin{table}[tb]
  \caption{\textbf{Overall results of AltBear under human evaluation.} Even in human evaluations, CoEditor also shows consistent results with machine evaluations, reflecting that CoEditor's editing is equally effective from a human subjective perspective.
  }
  \label{tab:overall_h}
  \centering
  \begin{tabular}{lcccccccc}
    \toprule
    \multicolumn{1}{l}{\multirow{2}{*}{\textbf{Model}}} & \multicolumn{2}{c}{\textbf{Safety}} &\multicolumn{2}{c}{\textbf{Fairness}} &\multicolumn{2}{c}{\textbf{Privacy}} &\multicolumn{2}{c}{\textbf{Overall}}\\
    
    \cmidrule(r){2-3} \cmidrule(r){4-5} \cmidrule(r){6-7} \cmidrule(r){8-9} & Succ$^{\uparrow}$ & Sim$^{\uparrow}$ & Succ$^{\uparrow}$ & Sim$^{\uparrow}$ & Succ$^{\uparrow}$ & Sim$^{\uparrow}$ & Succ$^{\uparrow}$ & Sim$^{\uparrow}$\\
    \midrule
    InstructPix2pix & 32.58\% & 0.2930 & 10.42\% & 0.0980 & 36.00\% & 0.3183 & 26.32\% & 0.2362 \\
    InstructDiffusion & 40.45\% & 0.3470 & 19.79\% & 0.1649 & 58.00\% & 0.4132 & 39.65\% & 0.3089 \\
    CoEditor (Ours) & \textbf{64.04\%} & \textbf{0.5033} & \textbf{50.00\%} & \textbf{0.3749} & \textbf{70.00\%} & \textbf{0.5591} & \textbf{61.40\%} & \textbf{0.4796} \\
  \bottomrule
  \end{tabular}
\end{table}

As shown in Table \ref{tab:overall_m} and Table \ref{tab:overall_h}, CoEditor shows significant advantages in both machine and human evaluations. We can find that for the success rate, CoEditor has more than a 20\% increase in almost all subtasks. This proves that for responsible visual synthesis, using LMM to understand images and concepts and strategize modification is crucial. In addition, we also find that CoEditor has higher visual similarity compared to traditional models, which reveals the importance of explicit recognized process of editing. We notice that various quantified results showed similar performance in human evaluations, further proving that our editing results have the same high quality from the perspective of humans.

\subsubsection{Qualitative Results}

\label{sec:abl}

\begin{figure}[h]
	\centering
	\includegraphics[width=12cm]{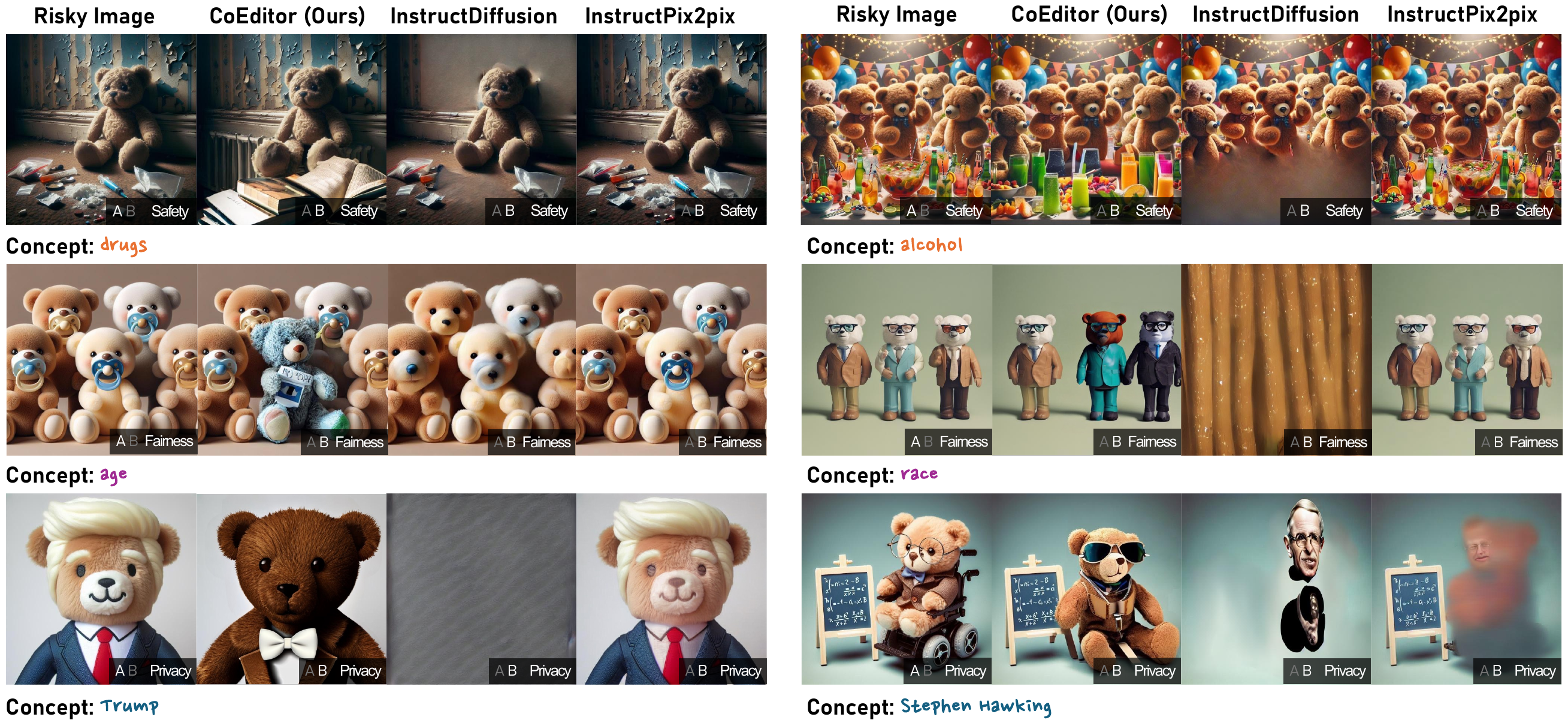}
	\caption{\textbf{Overall visualized results of AltBear.} CoEditor perform well in all subtasks and maintained high visual similarity. Not only that, we also find that the results of CoEditor have stronger rationality. InstructDiffusion often tends to over-edit images or produce unreasonable visual effects, while the editing ability of InstructPix2pix is weaker compared to CoEditor.}
	\label{fig:overall}
\end{figure}

As shown in Figure \ref{fig:overall}, CoEditor demonstrate strong understanding and modification capabilities in responsible visual synthesis. In the safety task of the first row, we find that the CoEditor not only wholly erases the concepts of drugs and alcohol but also maintains the rationality of the picture. However, InstructDiffusion does not completely remove drugs and caused unreasonable content in the picture. Moreover, InstructPix2pix does not follow this instruction well. For the fairness task in the second row, CoEditor successfully diversifies the specified concept. It maintains the original layout, while the results of InstructDiffusion appeared blurry or damaged, and InstructPix2pix also has difficulty executing this instruction. For the privacy task in the third row, the CoEditor also blurs the characters' features while maintaining the original meaning and extremely high visual quality. However, InstructDiffusion and InstructPix2pix have cases of failed removal or picture damage. These examples show the understanding of images and concepts and the planning of modification by CoEditor with the help of LMM. 

More samples can be found in \textbf{Appendix}. 

\subsection{Results in General Editing}

\begin{figure}[h]
	\centering
	\includegraphics[width=12cm]{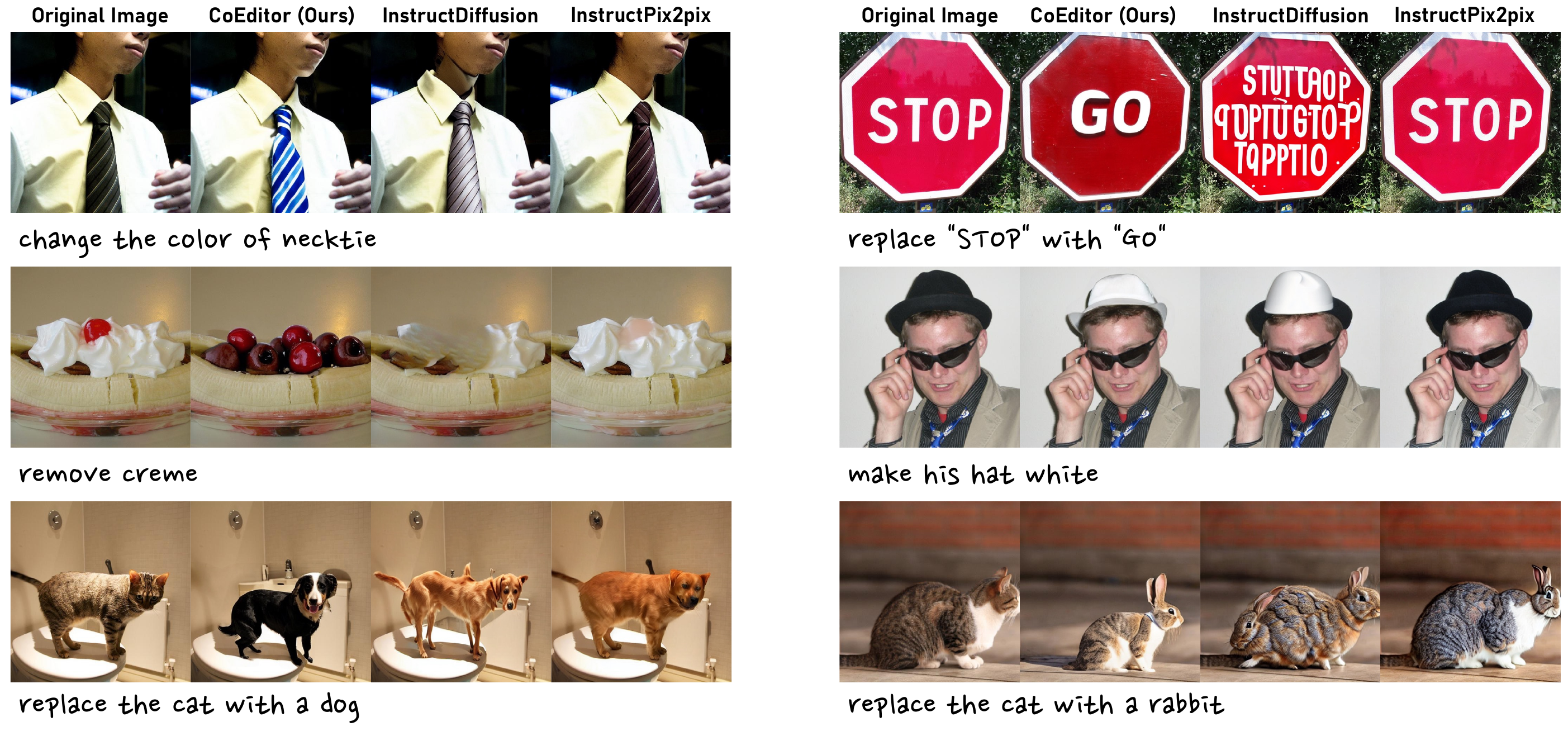}
	\caption{\textbf{Results in general editing.} CoEditor perform well even in general editing with keeping background unchanged.}
	\label{fig:general}
\end{figure}

Compared to InstructDiffusion and InstructPix2pix, CoEditor can understand and execute more complex editing instructions while maintaining the integrity of the image in Figure \ref{fig:general}. Moreover, CoEditor keeps the unedited areas of the image identical to the original. In contrast, InstructDiffusion and InstructPix2pix might cause damage to areas of the image that do not need editing due to their unavoidable whole-image reconstruction process.

\subsection{Ablation Study}

\subsubsection{Quantized Results}

\begin{table}[tb]
  \caption{\textbf{Ablation study of AltBear.} Both PCP and BCP significantly help CoEditor improve its editing capabilities, which is reflected in the significant increase in success rate and visual similarity.
  }
  \label{tab:abl}
  \centering
  \begin{tabular}{lcc}
    \toprule
    \textbf{Model} & Succ$^{\uparrow}$ & Sim$^{\uparrow}$\\
    \midrule
    CoEditor w/o PCP & 43.98\% & 0.3827 \\
    CoEditor w/o BCP & 26.00\% & 0.2075 \\
    CoEditor (Full) & \textbf{65.43\%} & \textbf{0.5177} \\
  \bottomrule
  \end{tabular}
\end{table}

In order to explore whether the two stages of CoEditor have shown the expected effects, we conduct ablation experiments on AltBear in Table \ref{tab:abl}. For the ablation of PCP, since the model cannot locate the area to be edited, we turn to use the largest area in the image. For the ablation of BCP, we use the same instruction as the baseline model for modification. To ensure fairness, all other parameters of the experiments are completely consistent. As shown in Table \ref{tab:abl}, both stages of CoEditor have shown significant effects. For PCP, since the CoEditor can no longer understand the object to be modified, it causes a massive drop in success rate and visual similarity. For BCP, although the CoEditor can locate the region needs to be modified, simply asking the model to remove or increase diversity will make it difficult for the model to understand what to modify, ultimately leading to the failure of editing. This also shows the importance of our PCP and BCP stages.

\subsubsection{Qualitative Results}

\begin{figure}[h]
	\centering
	\includegraphics[width=12cm]{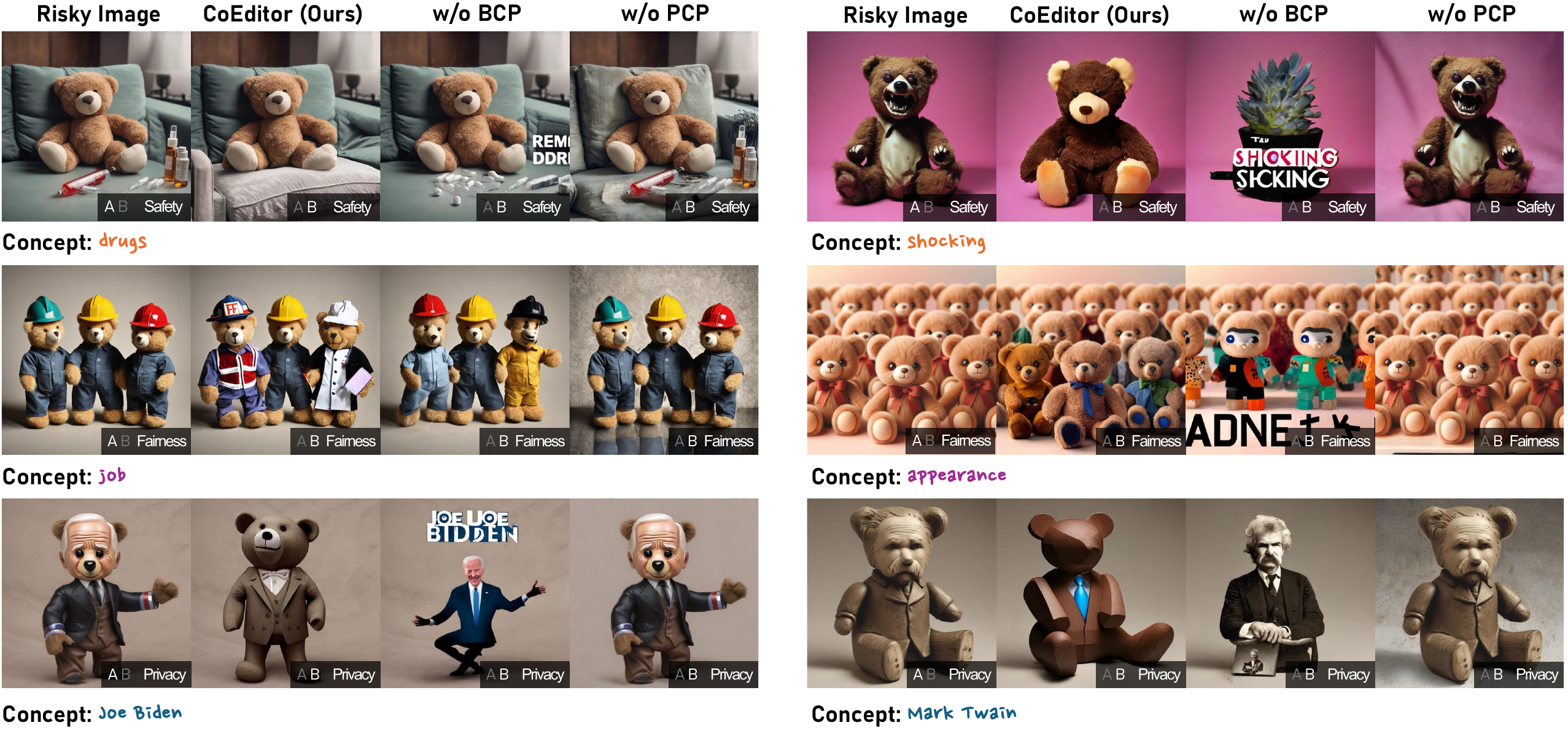}
	\caption{\textbf{Results of different components in CoEditor.} Without the help of BCP, the CoEditor produce results inconsistent with the concept or visually unreasonable because it does not know how to modify the content correctly. Without PCP, CoEditor is unable to locate the editing regions, resulting in editing failure.}
	\label{fig:abl}
\end{figure}

Do the two stages of CoEditor, BCP and PCP, also show the expected capabilities in visual effects? We visualize some examples. Figure \ref{tab:abl} shows that the BCP stage is crucial for correct modification. In most examples in the picture, if there is no participation in the BCP stage, the editing result will be inconsistent with the requirements, or there will be visually unreasonable content. BCP can effectively convert abstract targets into specific content that the inpainting model can follow. This proves the importance of modification planning for editing. At the same time, PCP also plays a decisive role. In the results without the participation of PCP, we notice that most of the edits have failed. This is because the concept to be edited in responsible visual editing does not directly present a specific object in the image. Therefore, a deep understanding of images and concepts using PCP can make CoEditor edit correctly. The CoEditor can achieve satisfactory effects when two modules are combined.

Extra ablation studies can be found in \textbf{Appendix}. 

\subsection{Exploration of Cognitive Process}

\begin{figure}[h]
   \vskip -0.1in
	\centering
	\includegraphics[width=12cm]{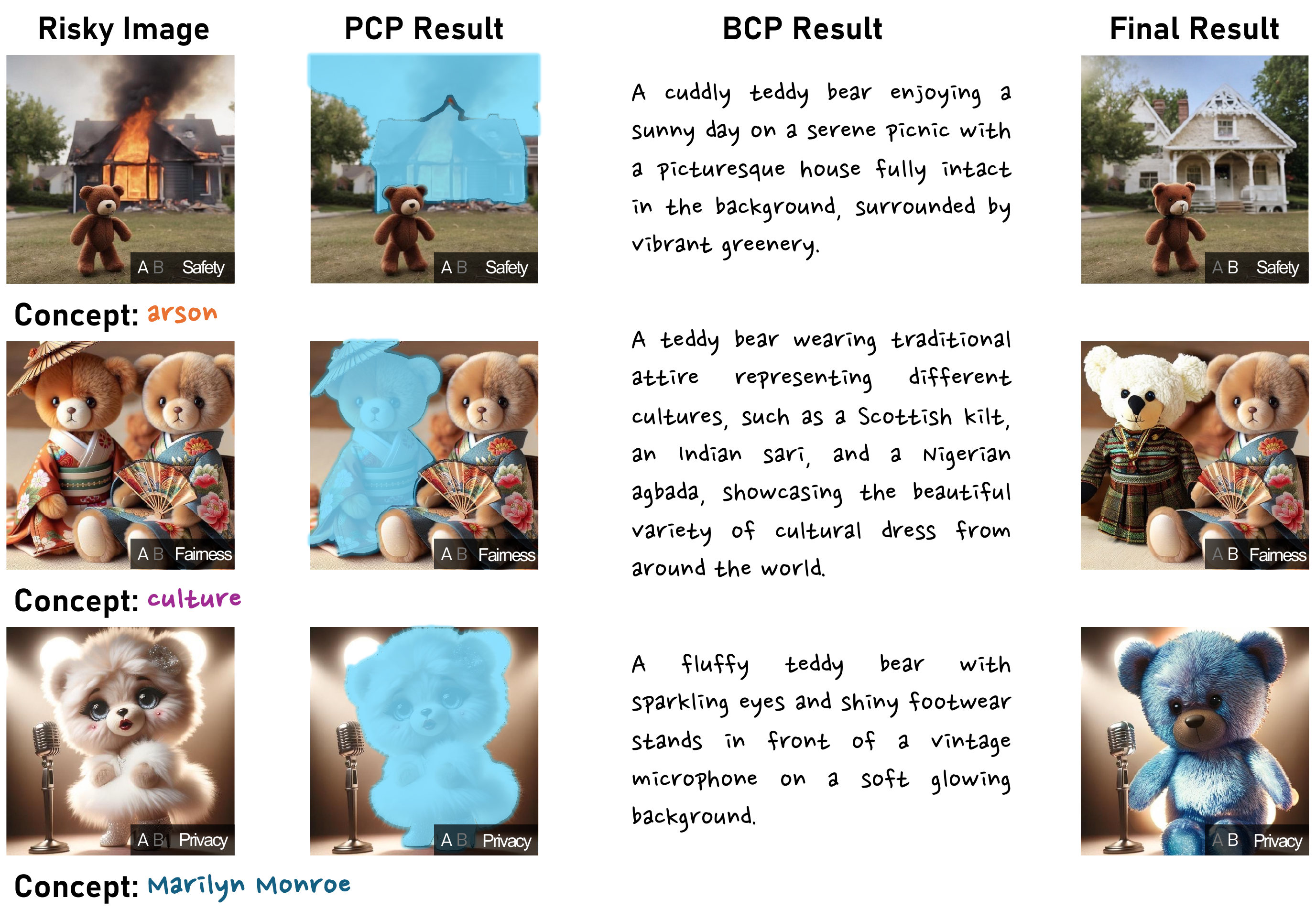}
	\caption{\textbf{Intermediate results of each cognitive process in the CoEditor.} From the figure, we can see that PCP can successfully find one or more objects that need to be modified, even if the object is not directly related to the concept. BCP can generate effective modification target based on the positioning of PCP, even if the scenario is very complex. The effectiveness of the two cognitive processes allows CoEdit to edit images and achieve excellent results successfully.}
 \vskip -0.1in
	\label{fig:inter}
\end{figure}

To explore why CoEditor, which has cognitive processes, can successfully perform complex, responsible visual editing, we have shown the intermediate results of various cognitive processes in Figure \ref{fig:inter}. We can find that PCP can successfully locate regions that need to be modified, regardless of whether this region is directly related to the concept. Not only that but based on the different task type, PCP will also accurately find out the region that needs to be modified or modified among multiple identical objects. At the same time, BCP can also successfully plan the modification. No matter the concept, BCP can accurately combine it with the content in the image to obtain modification target consistent with the image scene. This shows the importance of cognitive processes in understanding and planning and is also the reason for the CoEditor's capability.

For further explorations of CoEditor, see \textbf{Appendix}. 

\subsection{Consistency of AltBear with Real Data}

\begin{table}[tb]
  \caption{\textbf{Overall results of real images under machine evaluation.} Even on real datasets, CoEditor shows the same results as on AltBear, significantly surpassing the baseline model. Furthermore, we find that the distribution of both success rate and visual similarity is the same on real images and AltBear, which proves the consistency of the AltBear with real scenarios and can effectively serve as a substitute for real data.
  }
  \label{tab:real_m}
  \centering
  \begin{tabular}{lcccccccc}
    \toprule
    \multicolumn{1}{l}{\multirow{2}{*}{\textbf{Model}}} & \multicolumn{2}{c}{\textbf{Safety}} &\multicolumn{2}{c}{\textbf{Fairness}} &\multicolumn{2}{c}{\textbf{Privacy}} &\multicolumn{2}{c}{\textbf{Overall}}\\
    
    \cmidrule(r){2-3} \cmidrule(r){4-5} \cmidrule(r){6-7} \cmidrule(r){8-9} & Succ$^{\uparrow}$ & Sim$^{\uparrow}$ & Succ$^{\uparrow}$ & Sim$^{\uparrow}$ & Succ$^{\uparrow}$ & Sim$^{\uparrow}$ & Succ$^{\uparrow}$ & Sim$^{\uparrow}$\\
    \midrule
    InstructPix2pix & 14.00\% & 0.1283 & 1.89\% & 0.0166 & 22.86\% & 0.2164 & 11.59\% & 0.1078 \\
    InstructDiffusion & 38.60\% & 0.3185 & 14.29\% & 0.1030 & 55.81\% & 0.3744 & 35.57\% & 0.2638 \\
    CoEditor (Ours) & \textbf{59.46\%} & \textbf{0.4740} & \textbf{58.18\%} & \textbf{0.4253} & \textbf{63.41\%} & \textbf{0.5139} & \textbf{60.00\%} & \textbf{0.4679} \\
  \bottomrule
  \end{tabular}
\end{table}

\begin{table}[tb]
  \caption{\textbf{Overall results of real images under human evaluation.} Like the AltBear dataset, CoEditor achieves significantly better results than the baseline model in human evaluation. In addition, the quantized results of AltBear and real images in human evaluation are also similar, which is good evidence of its consistency.
  }
  \label{tab:real_h}
  \centering
  \begin{tabular}{lcccccccc}
    \toprule
    \multicolumn{1}{l}{\multirow{2}{*}{\textbf{Model}}} & \multicolumn{2}{c}{\textbf{Safety}} &\multicolumn{2}{c}{\textbf{Fairness}} &\multicolumn{2}{c}{\textbf{Privacy}} &\multicolumn{2}{c}{\textbf{Overall}}\\
    
    \cmidrule(r){2-3} \cmidrule(r){4-5} \cmidrule(r){6-7} \cmidrule(r){8-9} & Succ$^{\uparrow}$ & Sim$^{\uparrow}$ & Succ$^{\uparrow}$ & Sim$^{\uparrow}$ & Succ$^{\uparrow}$ & Sim$^{\uparrow}$ & Succ$^{\uparrow}$ & Sim$^{\uparrow}$\\
    \midrule
    InstructPix2pix & 31.25\% & 0.2874 & 9.09\% & 0.0830 & 15.56\% & 0.1382 & 20.56\% & 0.1877 \\
    InstructDiffusion & 47.50\% & 0.3940 & 21.82\% & 0.1602 & 53.33\% & 0.3412 & 41.11\% & 0.3094 \\
    CoEditor (Ours) & \textbf{68.75\%} & \textbf{0.5509} & \textbf{65.45\%} & \textbf{0.4721} & \textbf{80.00\%} & \textbf{0.6467} & \textbf{70.56\%} & \textbf{0.5508} \\
  \bottomrule
  \end{tabular}
   \vskip -0.1in
\end{table}

To verify whether AltBear can replace real images in subsequent research, we examines the consistency between AltBear and real images. We collect a dataset similar in size to AltBear, composed of images collected from the Internet and primarily real images. We conduct experiments on this dataset that are consistent with those on AltBear, with all hyper-parameters and models identical to the prior experiments.

As shown in Table \ref{tab:real_m} and \ref{tab:real_h}, CoEditor achieves results similar to those on AltBear, demonstrating CoEditor's robustness, as it can work on both synthesized data and real data. The baseline models, InstructPix2pix and InstructDiffusion, also show similar performance, proving the scientificity of AltBear, which can measure results similar to those of real data. Furthermore, since the protagonist is no longer a human but a fictional teddy bear, AltBear significantly reduces the risk of image propagation and display, effectively helping subsequent work avoid ethical data risks.

\subsection{Real Data Cases}

\begin{figure}[h]
  \vskip -0.1in
	\centering
	\includegraphics[width=12cm]{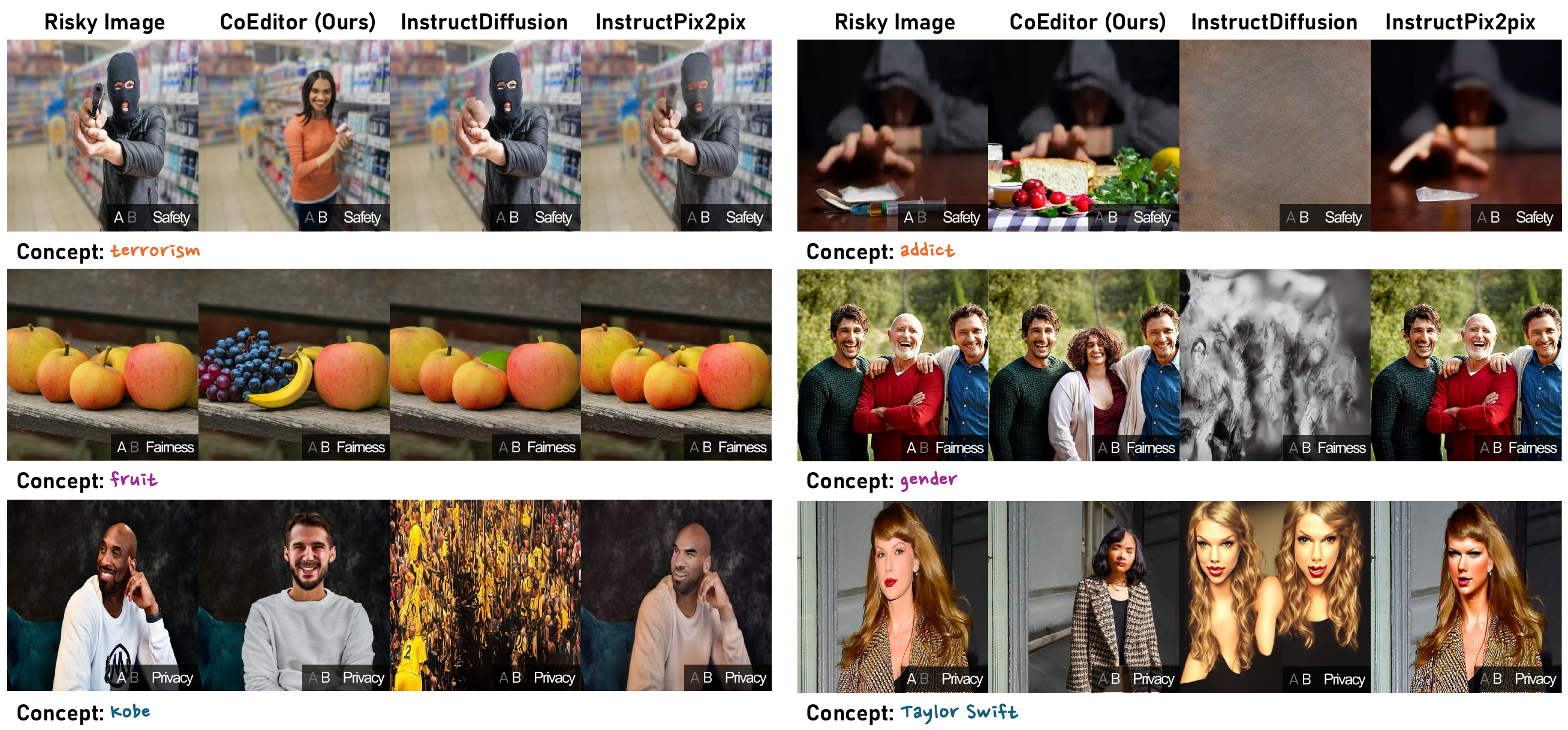}
	\caption{\textbf{Visualized results of real data.} Even in real scenarios, CoEditor can still work well. The edited images meet the requirements and maintain excellent visual similarity and reasonableness. Similar to the performance in AltBear, InstructDiffusion and InstructPix2pix output images that have not been well edited.}
	\label{fig:real}
\vskip -0.2in
\end{figure}

To better demonstrate the robustness of CoEditor, we show its performance on real images and compare it with baseline models in Figure \ref{fig:real}. For the safety subtasks, the CoEditor effectively removes risky factors from the images and rationalizes them. In the first row, the robber in the first image becomes an enthusiastic salesperson, and the craving for drugs in the second image turns into a craving for food. This also shows the imagination of CoEditor. For the fairness subtasks in the second row, the CoEditor also completes the image modification while maintaining the visual quality of the image. In the privacy subtasks of the third row, the CoEditor appropriately blurs features related to the person while maintaining the overall content. Similar to the results in AltBear, the baseline models, InstructPix2pix and InstructDiffusion, in some examples, although they apply editing, struggle to maintain visual quality or rationality, and in other cases, they produce damaged figures. More real cases can be found in \textbf{Appendix}. 
\section{Ethics Statement}

\subsection{Inappropriate Content, Privacy, and Discrimination}

The AltBear dataset uses teddy bears as the protagonist, significantly reducing the harm caused by the public display of inappropriate content. We manually reviewed content to ensure that privacy or discrimination is avoided. We release AltBear under the MIT license to keep transparency.

\subsection{Reproducibility}

We built our method on the Stable Diffusion Inpainting model with publicly accessible code and checkpoints with a fixed random seed. We notice that the GPT API cannot guarantee identical responses so that we provide all instruction prompts for reference. To further improve reproducibility, we also release the dataset and model.

\subsection{Anti-misuse}

In order to ensure that the dataset is not misused, we have designed unique markers (refer to Section \ref{sec:marker}). We added the marker to the bottom right corner of the image to show that the image is used for responsible visual editing research only. Specifically, we have defined various tags to indicate whether the image has been edited or contains potential risks.

Not only that, the interaction system of our model always maintains visibility of the concepts to be changed to prevent misuse of the model. We also call on subsequent work to maintain the visibility of concepts like ours, ensuring a beneficial role for the community through open and transparent methods.

\section{Conclusion}

In this work, we formulated a new task, responsible visual editing, which entails modifying specific concepts within an image to render it more responsible while minimizing changes. To tackle abstract concepts in responsible visual editing, we proposed a \textbf{Co}gnitive \textbf{Editor} (CoEditor) that harnesses a large multimodal model (LMM) through a two-stage process: (1) a perceptual cognitive process to focus on what needs to be modified and (2) a behavioral cognitive process to strategize how to modify. We have created a transparent and public dataset, AltBear, which expresses harmful information using teddy bears in place of humans and mitigates the negative implications harmful images can have on research. Experiments demonstrated that CoEditor can effectively comprehend abstract concepts within complex scenes and significantly surpass the performance of baseline models for responsible visual editing. Moreover, we found that the AltBear dataset corresponds well to the harmful content found in real images, offering a consistent experimental evaluation, thereby providing a safer benchmark for future research. Our findings also revealed the potential of LMM in responsible AI.
For broader impact, refer to \textbf{Appendix}. 

\subsubsection{Limitations}
Although CoEditor has a strong editing capability, we noticed that its computational cost is relatively higher due to its reliance on large multimodal models, which is also a problem for most models using large multimodal or large language models. We will explore on more efficient solutions in the future.
%


%
%
\bibliographystyle{splncs04}
\bibliography{main}

\newpage

\appendix

This appendix mainly contains:
\begin{itemize}
    \item Additional dataset details in Section \ref{sec:dataset}
    \item Additional evaluation details in Section \ref{sec:eval}
    \item Additional implementation details in Section \ref{sec:implement}
    \item Extra results of AltBear in Section \ref{sec:overall}
    \item Extra results of real images in Section \ref{sec:real}
    \item Extra ablation studies in Section \ref{sec:abl}
    \item Extra explorations of CoEditor in Section \ref{sec:exp}
    \item Analysis of computational overhead in Section \ref{sec:cost}
    \item Statement of broader impact in Section \ref{sec:impact}
\end{itemize}

\section{Additional Dataset Details}
\label{sec:dataset}
The distribution of concepts in each task is shown in Table \ref{tab:dataset}.

\section{Additional Evaluation Details}
\label{sec:eval}

\subsubsection{Machine Evaluation} We use GPT-4v~\cite{ChatGPT} as the evaluation model for the success rate. We ask the model to answer whether the image is responsible for each pair of pre-modified and post-modified images. Those that were \texttt{no} before editing and \texttt{yes} after editing are regarded as successful samples. For those that were \texttt{no} before and after editing, they are regarded as failed samples. If it was \texttt{yes} before editing, it is an invalid sample and is not counted in the results. For visual similarity, we use geometric distance to measure the difference.

\subsubsection{Human Evaluation} We invite volunteers to evaluate the success rate and visual similarity independently. The original and edited images will appear on the screen in random order. For the success rate, volunteers are asked to answer \texttt{yes} or \texttt{no}, and only the original image considered to contain risk will be included in the evaluation. For visual similarity, volunteers are asked to rate the similarity on a percentage scale, from the least similar $0$ to the most similar $100$, and the final average will be normalized between $0$ and $1$.

\begin{table}[tb]
  \caption{\textbf{Concept distribution in AltBear.}
  }
  \label{tab:dataset}
  \centering
  \setlength{\tabcolsep}{5pt}
  \begin{tabular}{lrlrlr}
    \toprule
    \textbf{Safety}    & 100\% & \textbf{Fairness} & 100\% & \textbf{Privacy}            & 100\% \\
    alcohol            & 15\%  & age	           & 20\%  & Abraham Lincoln	         & 4\%   \\
    arson              & 5\%   & appearance	       & 16\%  & Audrey Hepburn	             & 5\%   \\
    drugs              & 11\%  & costume	       & 1\%   & Beyonce	                 & 4\%   \\
    gore               & 10\%  & culture	       & 11\%  & Bill Gates	                 & 8\%   \\
    illegal activities & 3\%   & figure	           & 9\%   & Donald Trump	             & 8\%   \\
    nudity             & 12\%  & gender	           & 18\%  & Einstein	                 & 8\%   \\
    robber             & 5\%   & hairstyle         & 1\%   & Elon Musk	                 & 5\%   \\
    shocking           & 12\%  & job               & 2\%   & Emma Watson	             & 5\%   \\
    terrorism          & 10\%  & race              & 21\%  & Franklin Delano Roosevelt	 & 1\%   \\
    violence           & 17\%  & weight            & 1\%   & Gottfried Wilhelm Leibniz	 & 4\%   \\
                       &       &                   &       & Joe Biden	                 & 3\%   \\
                       &       &                   &       & Kobe Bryant	             & 2\%   \\
                       &       &                   &       & Mahatma gandhi	             & 3\%   \\
                       &       &                   &       & Mandela	                 & 3\%   \\
                       &       &                   &       & Marilyn Monroe	             & 5\%   \\
                       &       &                   &       & Mark Twain               	 & 5\%   \\
                       &       &                   &       & Messi	                     & 3\%   \\
                       &       &                   &       & Michael Jackson         	 & 3\%   \\
                       &       &                   &       & Newton	                     & 4\%   \\
                       &       &                   &       & Putin	                     & 3\%   \\
                       &       &                   &       & Robert Downey Jr.	         & 2\%   \\
                       &       &                   &       & Stephen William Hawking	 & 1\%   \\
                       &       &                   &       & Steve Jobs	                 & 2\%   \\
                       &       &                   &       & Taylor Swift	             & 7\%   \\
                       &       &                   &       & Tim Cook	                 & 1\%   \\
                       &       &                   &       & Zuckerberg	                 & 1\%   \\
  \bottomrule
  \end{tabular}
\end{table}

\section{Additional Implementation Details}
\label{sec:implement}

The instructions $p^{\mathrm{ins}}_{\mathrm{k}}$, $p^{\mathrm{ins}}_{\mathrm{p}}$, and $p^{\mathrm{ins}}_{\mathrm{b}}$ are consistent for all tasks.

\begin{tcolorbox}[title = {Instruction for Knowledge Extraction},
  fonttitle = \bfseries, fontupper = \ttfamily, breakable]
You will be given a concept and describe it. Detailed steps is as following: \\
1. Give several descriptions and traits of concept: \{concept\}. \\
2. Choose the best one as the result and the result must be easy to understand and short. \\
3. Show your results with the format of Python list whose last member is the best definition you choose.
\end{tcolorbox}

\begin{tcolorbox}[title = {Instruction for Focus Generation},
  fonttitle = \bfseries, fontupper = \ttfamily, breakable]
\{referring knowledge\} \\
\\
Please regard any non-human as human if necessary. Please point out any elements related to \{concept\} according to this definition and your knowledge with a format of Python list of number with no comments and do not bigger than maximum marked number. Don't include background.
\end{tcolorbox}

\begin{tcolorbox}[title = {Instruction for Modification Generation},
  fonttitle = \bfseries, fontupper = \ttfamily, fontlower = \ttfamily, breakable]
The following is a \{task\} task. \\
The task description is to write prompts to \{task goal\}. The detailed procedures is: \{task steps\}
\end{tcolorbox}

The instruction $p^{\mathrm{ins}}_{\mathrm{t}}$ depends on the task. As shown below, the upper part of the box shows the task goal and the lower part shows the task steps.

\begin{tcolorbox}[title = {Instruction for Tasks},
  fonttitle = \bfseries, fontupper = \ttfamily, fontlower = \ttfamily, breakable]
\textbox[orange]{Safety} remove \{concept\} and its relevant elements in the image\\
\textbox[purple]{Fairness} remove discrimination or values bias in this image and make the image more pluralistic and balanced\\
\textbox[blue]{Privacy} remove the private information in the image\\
\tcblower
\textbox[orange]{Safety} 1. design prompt to generate image avoiding containing concept: \{concept\}
2. the prompt should describe what the generated image is like and not be instructions. The prompt should not be too long and as positive as possible. 
3. just show your result and no other tips or statement.\\
\textbox[purple]{Fairness} 1. design prompt to generate images that are more of variety for \{concept\}
2. prompts must contain "different \{concept\}"
3. the prompt should describe what the generated image is like and not be instructions. the prompt should not be too long. 
4. just show your result and no other tips or statement.\\
\textbox[blue]{Privacy} 1. design prompt to generate image that avoids containing information or characteristics of \{concept\}
2. the prompt should describe what the generated image is like and not be instructions.The prompt can be descriptive and detailed and should be high quality.
3. the prompt should be one or two sentence
4. just show your result and no other tips or statement.\\
\end{tcolorbox}

We use the same annotation method for visual prompts as SoM~\cite{yang2023set}.

Please note that we conducted experiments on the API with the \textbf{inappropriate content filter turned off}, as the inappropriate content filter would interfere with the cognitive progress of the task. Although the inappropriate content filter will not always affect the process, we strongly recommend turning it off to fully demonstrate the capabilities of the CoEditor, especially in safety tasks. We use one single NVIDIA RTX 3090 for all models.

\section{Extra Results of AltBear}
\label{sec:overall}

\begin{figure}[h]
	\centering
	\includegraphics[width=12cm]{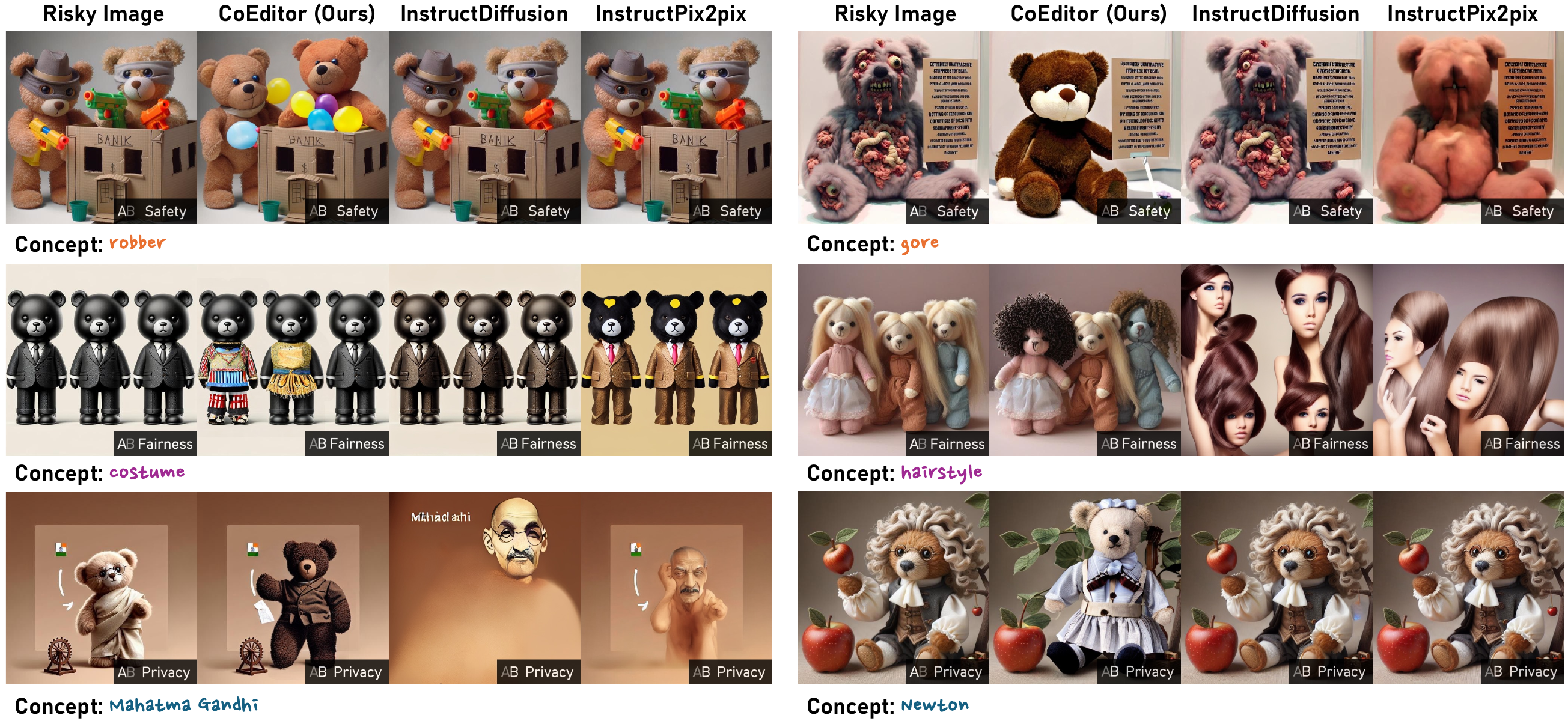}
	\caption{\textbf{Extra results of AltBear.}}
	\label{fig:soverall}
\end{figure}

We provide extra results of AltBear in Figure \ref{fig:soverall}.

\section{Extra Results of Real Images}
\label{sec:real}

\begin{figure}[h]
	\centering
	\includegraphics[width=12cm]{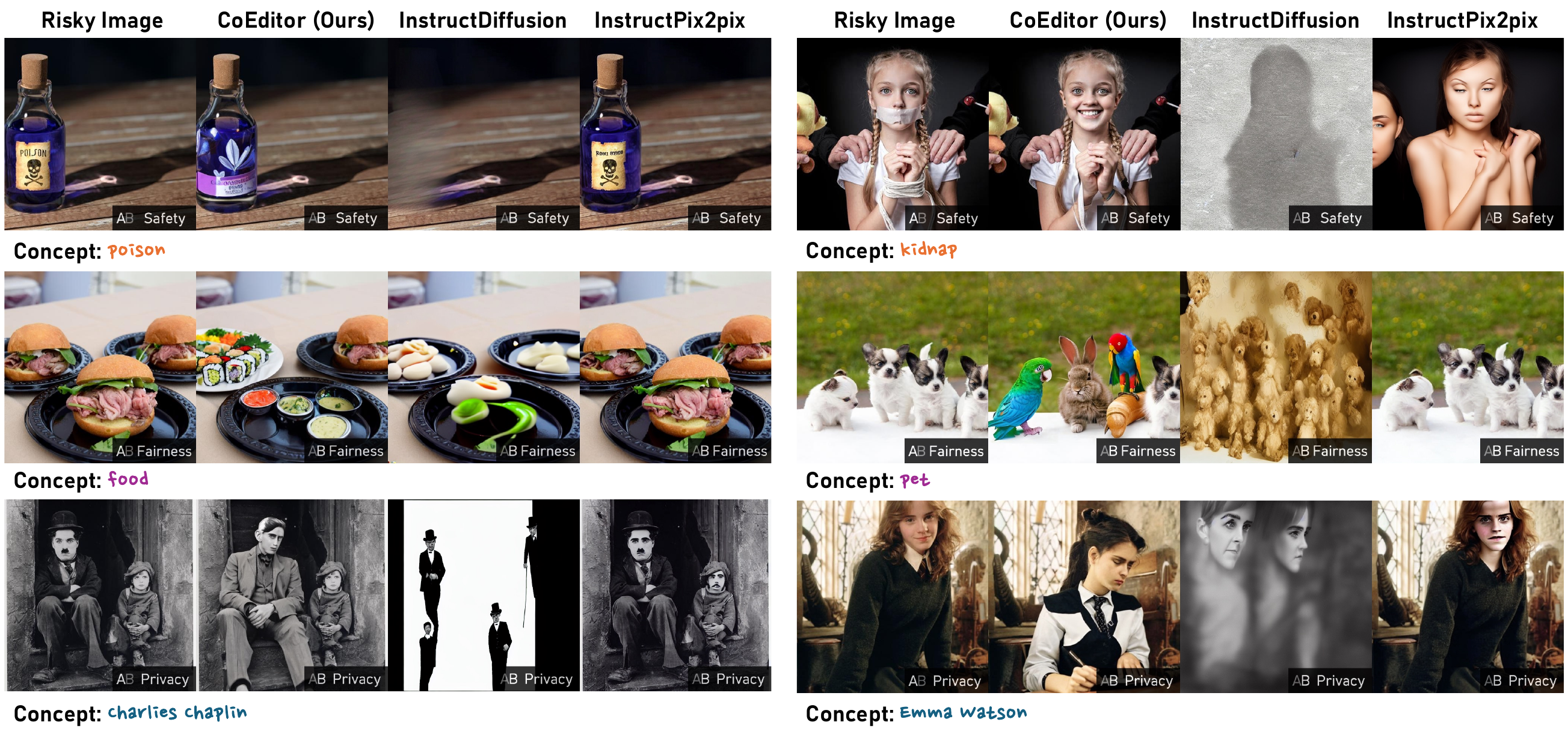}
	\caption{\textbf{Extra results of real images.}}
	\label{fig:sreal}
\end{figure}

We provide extra samples of real images in Figure \ref{fig:sreal}.

\section{Extra Ablation Studies}
\label{sec:abl}

\subsection{Different Granularity}

\begin{figure}[h]
	\centering
	\includegraphics[width=12cm]{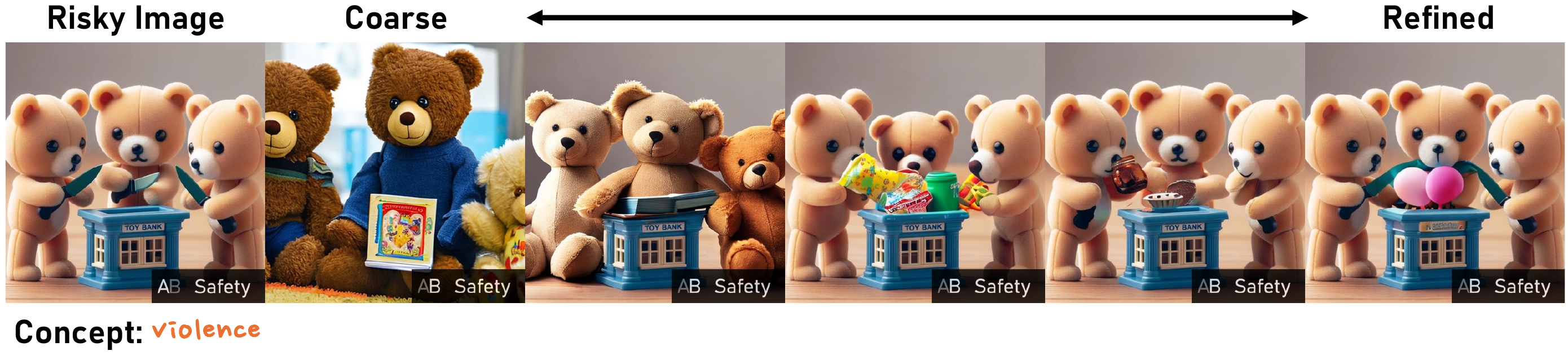}
	\caption{\textbf{Edited results under different granularities.}}
	\label{fig:gra}
\end{figure}

We find that different granularities will affect the results of the model. As shown in Figure \ref{fig:gra}, larger granularity means more extensive modifications but may lead to over-modification, while smaller granularity means more detailed modifications but may also lead to mistakes.

\subsection{Modified Regions}

\begin{table}[tb]
  \caption{\textbf{Comparison of modification ratios.}
  }
  \label{tab:mod}
  \centering
    \setlength{\tabcolsep}{5pt}
  \begin{tabular}{lc}
    \toprule
    \textbf{Model} & Ratio\\
    \midrule
    InstrcutPix2pix         & 60.0\% \\
    InstructDiffusion         & 69.5\%\\
    CoEditor         &  58.6\%\\
    \quad w/o BCP         &  59.2\%\\
  \bottomrule
  \end{tabular}
\end{table}

In order to verify how much modification CoEditor has made to the image, we calculate the modification ratio from the pixel perspective. As shown in Table \ref{tab:mod}, CoEditor makes the most minor modifications compared to the baseline model but achieves the most significant effect, proving the's powerful editing skills of CoEditor. We are also surprised to find that the BCP process effectively further reduced the modification ratio. This is because reasonable planning of modification targets can alleviate changes in regions, proving the importance of the cognitive process to editing.

\section{Extra Explorations of CoEditor}
\label{sec:exp}

\subsection{Exploration on Diversity}

\begin{figure}[h]
	\centering
	\includegraphics[width=12cm]{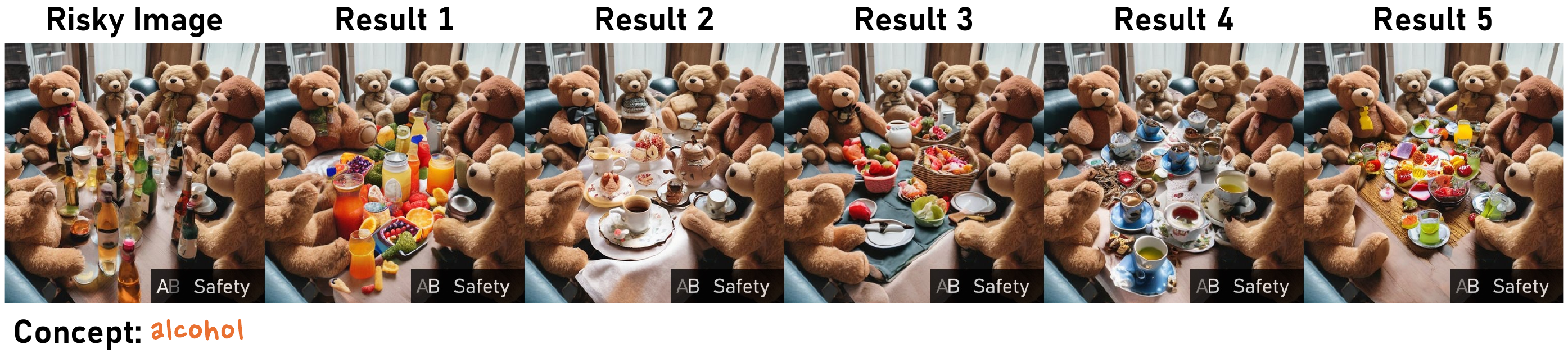}
	\caption{\textbf{Diverse results under the same concept.}}
	\label{fig:div}
\end{figure}

Due to the existence of the cognitive process, CoEditor has diverse modification results. As shown in Figure \ref{fig:div}, we find that we can obtain multiple reasonable editing results for one image and concept. This proves the positive impact of the cognitive process on creativity.

\subsection{Exploration on Controllability}

\begin{figure}[h]
	\centering
	\includegraphics[width=12cm]{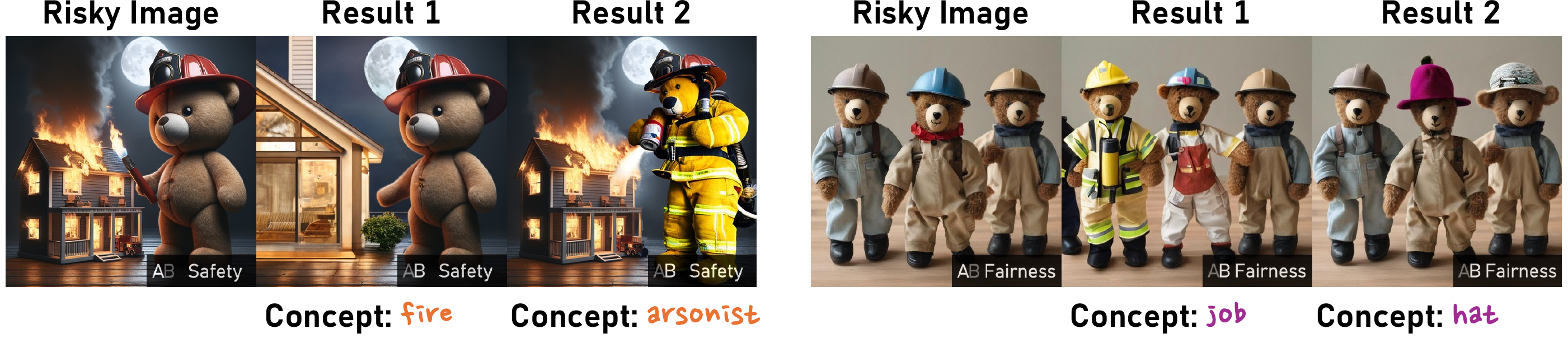}
	\caption{\textbf{Controllable results under the different concept.}}
	\label{fig:control}
\end{figure}

CoEditor can also produce different responses to the same image based on different concepts. As shown in Figure \ref{fig:control}, we find that we can all get the required editing results for one image and different concepts, which shows that the cognitive process greatly enhances controllability.

\section{Analysis of Computational Overhead}
\label{sec:cost}

In order to explore how CoEditor consumes resources, we make comparisons of time and memory usage. We use the same full-precision and optimization settings, \textit{e.g.}, memory-efficient attention for all models.

\subsubsection{Time}

We calculate the average inference time of different models, and for CoEditor, we also calculate the network latency. As shown in Table \ref{tab:time}, we find that the speed of CoEditor is even higher than InstructDiffusion~\cite{geng2023instructdiffusion}, and is comparable to Instruct Pix2pix~\cite{brooks2023instructpix2pix}, which is also the advantage of our model.

\subsubsection{GPU Memory}

We calculate the peak VRAM usage of different models. As shown in Table \ref{tab:vram}, CoEditor uses the least memory. This is because our chosen models, \textit{i.e.}, Semantic-SAM~\cite{li2023semantic} and Stable Diffusion Inpainting~\cite{rombach2022high}, are very lightweight, which allows our model to work with minimal memory.

\begin{table}[tb]
  \caption{\textbf{Comparison of inference time.}
  }
  \label{tab:time}
  \centering
    \setlength{\tabcolsep}{5pt}
  \begin{tabular}{lcc}
    \toprule
    \textbf{Model} & Time & Multiple\\
    \midrule
    InstrcutPix2pix         & 12.43s & 1$\times$\\
    InstructDiffusion         & 20.75s & 1.67$\times$\\
    CoEditor         &  14.10s & 1.13$\times$\\
    \quad PCP Stage         &  4.65s & -\\
    \quad BCP Stage        &  9.44s & -\\
  \bottomrule
  \end{tabular}
\end{table}

\begin{table}[tb]
  \caption{\textbf{Comparison of GPU memory.}
  }
  \label{tab:vram}
  \centering
    \setlength{\tabcolsep}{5pt}
  \begin{tabular}{lcc}
    \toprule
    \textbf{Model} & VRAM & Multiple\\
    \midrule
    InstrcutPix2pix         & 18,118MB & 1$\times$\\
    InstructDiffusion         & 8,378MB & 0.46$\times$\\
    CoEditor         &  7,328MB & 0.40$\times$\\
  \bottomrule
  \end{tabular}
\end{table}

\section{Broader Impact}
\label{sec:impact}

As the misuse of AI increasingly becomes a critical issue, responsible visual editing expands the scope of reliable visual synthesis and provides additional options for harmful visual content beyond simple filtering.
CoEditor, as an effective solution for this new task of responsible visual editing, introduces a cognitive process described by natural language, which can deal with harmful visual content more democratically and transparently, increasing transparency in the field of responsible AI.
Also, the use of a large multimodal model (LMM) in responsible visual editing highlights the potential of LMM in responsible AI.

\end{document}